\documentclass{article}
\usepackage{spconf,amsmath,epsfig}
\usepackage{times}  % DO NOT CHANGE THIS
\usepackage{helvet} % DO NOT CHANGE THIS
\usepackage{courier}  % DO NOT CHANGE THIS
\usepackage[hyphens]{url}  % DO NOT CHANGE THIS
\usepackage{graphicx} % DO NOT CHANGE THIS

\usepackage{times}
\usepackage{epsfig}
\usepackage{graphicx}
\usepackage{amsmath}
\usepackage{amssymb}
\usepackage{array}
\usepackage{cases}
\usepackage{color}
\usepackage{float}
\usepackage{amsmath}
\usepackage{multirow}
\usepackage{setspace}
\usepackage{txfonts}
\usepackage{algpseudocode}
\usepackage{booktabs} 
\usepackage{subfigure}
\usepackage{caption}
\usepackage{algorithm}
\usepackage{algpseudocode}
\usepackage{soul}
\usepackage{pifont}
\usepackage{times}
\usepackage{epsfig}
\usepackage{graphicx}
\usepackage{amssymb}
\usepackage{subfigure}
\usepackage{pifont}
\usepackage{color,xcolor,colortbl}
\usepackage{array}
\usepackage{url}
\usepackage{CJK}
\usepackage{makecell}
 \usepackage{arydshln}
\usepackage{array}     % and/or
\usepackage{longtable} % and/or
\usepackage{colortab}  % or
\usepackage{dcolumn}
\usepackage{array}
\usepackage{setspace}
\usepackage{diagbox}
\usepackage[numbers,sort]{natbib}

\newcolumntype{I}{!{\vrule width 1.2pt}}
\newlength\savedwidth

\let\OLDthebibliography\thebibliography
\renewcommand\thebibliography[1]{
  \OLDthebibliography{#1}
  \setlength{\parskip}{0pt}
  \setlength{\itemsep}{0pt plus 0.3ex}
}

\begin{document}\sloppy

% Example definitions.
% --------------------
\def\x{{\mathbf x}}
\def\L{{\cal L}}

% Title.
% ------
\title{Dilated-Scale-Aware Attention ConvNet for Multi-Class Object Counting}
%
% Single address.
% ---------------
\renewcommand{\thefootnote}{\fnsymbol{footnote}}
\name{Wei Xu$^1$\textsuperscript{$*$},
    Dingkang Liang$^2$\textsuperscript{$*$},
    Yixiao Zheng$^1$,
    Zhanyu Ma$^1$\textsuperscript{$\dagger$} 
    }

\address{\textsuperscript{1}Beijing University of Posts and Telecommunications, \textsuperscript{2}Huazhong University of Science and Technology\\
% weixu2020@bupt.edu.cn, dkliang@hust.edu.cn, zhengyixiao@bupt.edu.cn, mazhanyu@bupt.edu.cn
}

\maketitle
\protect \renewcommand{\thefootnote}{\fnsymbol{footnote}}
\footnotetext[1]{Equal contribution.} 
\footnotetext[2]{Corresponding author.}

\begin{abstract}
Object counting aims to estimate the number of objects in images. The leading counting approaches focus on single-category counting task and achieve impressive performance. Note that there are multiple categories of objects in real scenes. Multi-class object counting expands the scope of application of object counting task. The multi-target detection task can achieve multi-class object counting in some scenarios. However, it requires the dataset annotated with bounding boxes. Compared with the point annotations in mainstream object counting issues, the coordinate box-level annotations are more difficult to obtain. In this paper, we propose a simple yet efficient counting network based on point-level annotations. Specifically, we first change the traditional output channel from one to the number of categories to achieve multi-class counting. Since all categories of objects use the same feature extractor in our proposed framework, their features will interfere mutually in the shared feature space. We further design a multi-mask structure to suppress harmful interaction among objects. Extensive experiments on the challenging benchmarks illustrate that the proposed method achieves state-of-the-art counting performance. 
% \textit{The code will available at https://github.com/xuweigogogo/DSACA
\end{abstract}
%  

% \vspace{-5pt}
\section{Introduction}
\label{sec:intro}
Object counting is a prevalent task in the computer vision community, playing an essential role in a wide range of applications (\textit{e.g.,} crowd analysis, urban management, and wildlife conservation). Most counting methods count the number of a single class, such as crowd counting~\cite{gao2020nwpu}, cell counting~\cite{falk2019u}, and vehicle counting~\cite{guerrero2015extremely}, which can be grouped as single-class object counting methods. Specifically, the single-class object counting methods  require re-training when changing the category to be counted.

\begin{figure}[!t]
\centering
\resizebox{0.47\textwidth}{!}{
    \includegraphics{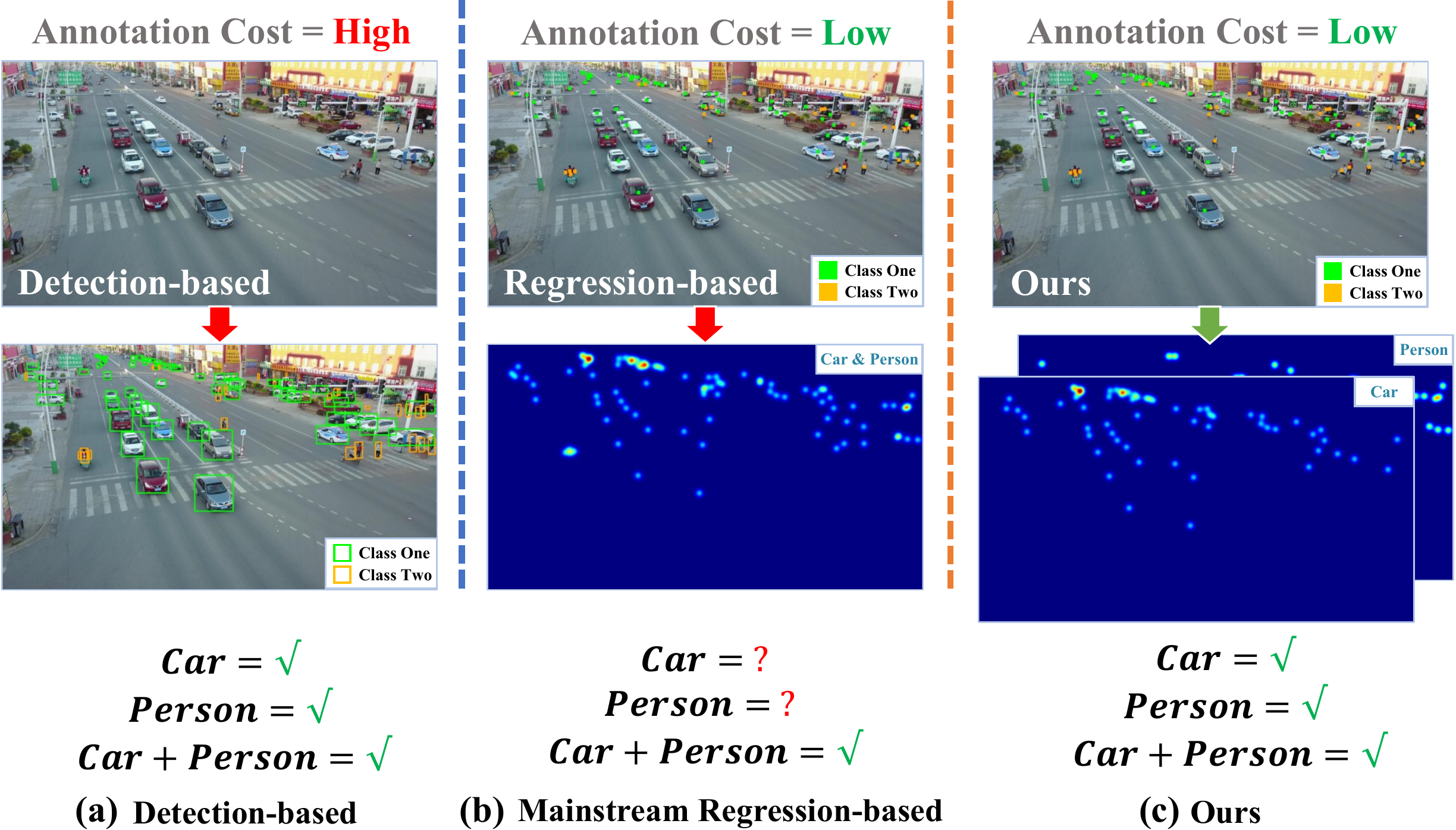}
    } 
\vspace{-3pt}
\caption{Comparisons of the proposed method with regression-based and detection-based object counting methods. (a) The detection-based methods rely on box-level annotations, but obtaining the box-level annotations in dense scenes is an expensive and labor-intensive process. (b) The mainstream regression-based methods are able to obtain the total count of all categories, but it's difficult to distinguish the number of each category. (c) The proposed method estimates density maps for each category without expensive box-level annotations.}
\centering
\label{fig:intro}
\vspace{-15pt}
\end{figure}
 
Few methods focus on counting multiple categories with a single model, and we define the problem as \textit{multi-class object counting}. Multi-class object counting is more practical in real-world applications. For instance, it is necessary to count the appeared persons and cars in an image to assess the degree of traffic congestion during rush hour.

% The detection-based methods~\cite{girshick2015fast, liu2016ssd,redmon2016you}, which can be implemented for multi-class object counting task by casting the problem as detecting each instance of image and obtain the number of each category by counting the bounding boxes, as shown in Fig.~\ref{fig:intro}(a).
Detection-based methods~\cite{ren2016faster, liu2016ssd,redmon2016you} enable multi-class object counting tasks by counting the number of bounding boxes in each category, as shown in Fig.~\ref{fig:intro}(a). However, the detection-based methods rely on bounding box annotations, which is impractical due to annotating the bounding boxes of each instance for training images can be a significant challenge in dense scenes. Thus, the mainstream counting datasets only provide point-level annotations, causing the detectors untrainable. 

The regression-based methods~\cite{xu2019learn, yang2020reverse, zhang2016single, guerrero2015extremely}, which only rely on point-level annotations, have achieved significant successes by adopting the density maps. However, the widely used density map is obtained by overlaying a series of Gaussian Blobs. It makes the category indistinguishable so that the current regression-based counting methods hardly implement multi-class object counting,  shown in Fig.~\ref{fig:intro}(b). 

To solve the problems mentioned above, we propose the Dilated-Scale-Aware Category-Attention ConvNet (DSACA) to achieve multi-class object counting, just relying on point-level annotations. DSACA learns multiple density maps for multi-class object counting instead of a single density map, as shown in Fig.~\ref{fig:intro}(c). Specifically, DSACA, adopting a VGG16~\cite{simonyan2015very} as the backbone, consists of two crucial modules named Dilated-Scale-Aware Module (DSAM) and Category-Attention Module (CAM), respectively. DSAM fuses the stage$\_3$, stage$\_4$, and stage$\_5$ feature maps from the backbone as the input and utilizes diverse dilated rates to capture the intra-class visual responses of big objects and tiny objects simultaneously. Sharing the same backbone, the density maps will interfere with each other, which inspires us to design the CAM to provide high-quality class-isolated density maps.
The main contributions of our method are summarized as follows:
\begin{enumerate}
\item To the best of our knowledge, this is the first attempt to implement the multi-class object counting task based on the point-level annotations in the published literature.
\item We propose a novel multi-class object counting method named DSACA consisting of DSAM and CAM. The DSAM effectively captures the multi-scale information in the intra-class scene, and the CAM shows strong adaptability to resist inter-class interference among density maps.
\item Extensive experiments demonstrate that our approach achieves state-of-the-art performance on two challenging datasets.
\end{enumerate}

\begin{figure*}[t]
\centering
\resizebox{0.94\textwidth}{!}{
    \includegraphics{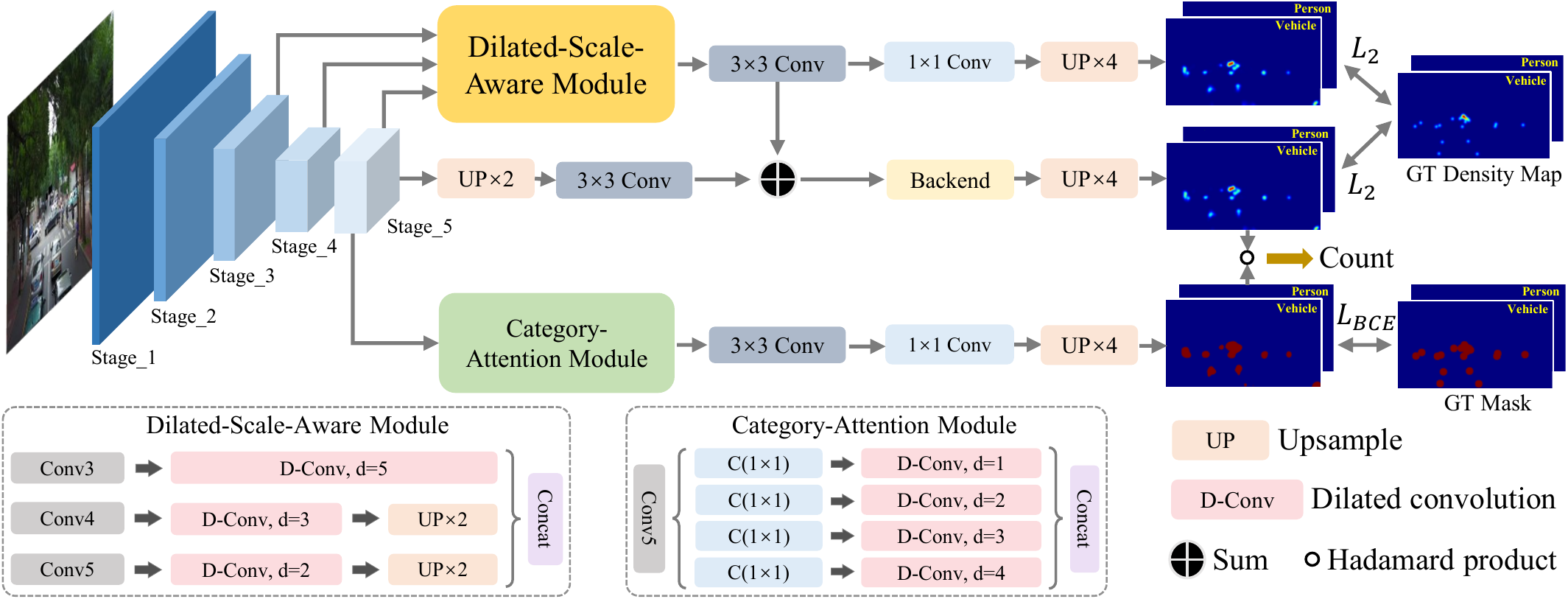}
    }
\caption{The pipeline of our method. The proposed approach fuse multi-scale features via Dilated-Scale-Aware Module and suppress harmful inter-class responses via Category-Attention Module.}
\centering
\label{fig:pipeline}
\vspace{-13pt}
\end{figure*}

% \begin{figure}[t]
% \centering
% \resizebox{0.47\textwidth}{!}{
%     \includegraphics{Scale-Aware Module-cropped.pdf}
%     }
% \caption{The structure of Dilated-Scale-Aware Module.}
% \centering
% \label{fig:Scale-Aware Module-cropped}
% \vspace{-10pt}
% \end{figure}

% \begin{figure}[t]
% \centering
% \resizebox{0.47\textwidth}{!}{
%     \includegraphics{module_v2-cropped.pdf}
%     }
% \caption{******}
% \centering
% \label{fig:module}
% \end{figure}

% \vspace{-5pt}
\section{Related works}
The mainstream object counting methods can be categorized into detection-based methods and regression-based methods.

% \vspace{-5pt}
\subsection{Detection-based methods}
% \vspace{-5pt}
Early traditional methods capture the handcrafted features (\textup{e.g.,} Haar\cite{papageorgiou1998general} and HOG\cite{dalal2005histograms}) to discriminate the person or non-person. These traditional methods achieve limited performance due to heavy occlusions, scale variations, and perspective distortions. Recently, the detection-based methods~\cite{ren2016faster, zhou2019objects, liu2016ssd, redmon2016you} usually utilize deep object detectors to locate each instance since the advances of deep learning. Specifically, the pioneer RCNN~\cite{girshick2014rich} extracts region proposals from an image and utilizes a convolutional neural network to classify each region of interest (ROI) independently.  SSD~\cite{liu2016ssd} is a one-stage detector, which introduces the multi-reference and multi-resolution detection fusion techniques. These deep-learning-based detectors rely on bounding-boxes annotations, but obtaining bounding-boxes annotations is an expensive and laborious process in the dense scenes.

% \vspace{-5pt}
\subsection{Regression-based methods}
% \vspace{-10pt}
Recently, various regression-based object counting methods~\cite{ma2019bayesian, zhang2016single, xu2019learn, liu2019adcrowdnet} regress a density map through different techniques such as multi-scale network~\cite{xu2019learn, Jiang_2019_CVPR}, attention mechanism~\cite{bai2020adaptive,liu2019adcrowdnet,zhang2019attentional}, and perspective information~\cite{yan2019perspective,liu2019context,yang2020reverse}, achieving remarkable progress. Specifically, Jiang \textit{et al.}~\cite{Jiang_2019_CVPR} introduce a novel Trellis style encoder-decoder, which effectively fuse multi-scale feature maps in the encoder. Xu \textit{et al.}~\cite{xu2019learn} propose the Learning to Scale Module (L2SM) to automatically scale different region into similar densities, reducing the pattern shift. Zhang \textit{et al.}~\cite{zhang2019attentional} propose the Attentional Neural Field (ANF), which combines the conditional random ﬁelds with non-local attention mechanisms to capture multi-scale features and long-range dependencies. Yang \textit{et al.}~\cite{yang2020reverse} propose a reverse perspective network, which is designed to diminish the scale variations in an unsupervised way. Liu \textit{et al.}~\cite{liu2019context} utilize a auxiliary branch to integrate perspective maps into the density maps. 

Regression-based methods usually adopt MSE loss as their loss function. However, only use the MSE loss function will cause some problems, such as blur eﬀect, neglecting the local consistency, and losing position information. Therefore, designing an appropriate loss function is also an essential researching direction in the training stage, which can promote the object counting ability. BL~\cite{ma2019bayesian} regards the density map as a probability map, computing the probability of each pixel. SPANet~\cite{cheng2019learning} proposes Maximum Excess over Pixels (MEP) loss by ﬁnding the pixel-level subregion with the highest discrepancy with ground truth.

In general, these regression-based methods only focus on single class object counting (\textit{e.g.,} crowd counting), which limits the applications in the real-world.
%All of these regression-based method 

\vspace{-10pt}
\section{Proposed method}
The overview pipeline of our method is shown in Fig.~\ref{fig:pipeline}. The proposed method mainly contains two crucial components: the Dilated-Scale-Aware Module (DSAM) and Category-Attention Module (CAM). We utilize the pre-trained VGG-16 as the backbone and discard the pooling layer between the stage$\_4$ and stage$\_5$ to maintain the spatial size, which follows previous works~\cite{bai2020adaptive, yang2020reverse, xu2019learn}. The backend utilizes $3\times3$ convolutional layers and a $1\times1$ layer to generate the density maps. The DSAM is designed to capture the visual context cues by extracting features at different scales. The CAM focus on the inter-class feature decorrelation, which can effectively reduce the bad interference among density maps belonging to different categories.

% The proposed architecture is shown in Fig.~\ref{fig:pipeline}. It mainly consists of two modules: Scale-Aware Module (SAM), Category-Attention Module (CAM). We first adopt VGG16~\cite{simonyan2015very} without the last two pooling layers as the backbone, from which SAM and CAM obtain different levels of features. Based on the last 3 stages, SAM fuses the more granular visual information extracted from feature maps at specific stages of the convolutional layers with different dilation scales. To estimate the density maps for each class further, the scale-aware feature is then summarized with the feature maps captured from the frontend and fused  by the backend structure. Specifically, the pixels within a specific range around the annotations should be given greater attention. So, to obtain more effective density maps, CAM generate hard attention maps and matrix point multiplication with previously estimated density maps to get the final maps. The proposed modules fully capture the attention information in terms of feature scale, instance size, instance location and so on aspects to improve the counting accuracy.

% We first introduce SAM and CAM in Section 3.1, Section 3.2. Then in Section 3.3, we introduce the loss functions that constrain these two modules.

\vspace{-10pt}
\subsection{Multi-Class object counting problem definition}\label{section3.1}
The current single-class object counting methods apply the Gaussian kernels on point annotations to generate density maps as ground truth and make predictions about the number of a single class. To realize the multi-class object counting, we generate individual density maps for each category in an image. To be specific, given an image containing $N$ classes of objects, the ground-truth density map $D$ can be illustrated as:

\begin{equation}
D=\left\{
\begin{matrix}
\sum_{i=1}^{C_1}\delta  ( P^{1}-P_{i}^{1}  )\ast G_{\sigma^{1} } ( P^{1} )
\\
\vdots
\\
\sum_{i=1}^{C_k}\delta  ( P^{k}-P_{i}^{k}  )\ast G_{\sigma^{k} } ( P^{k} ) ,
\\
\vdots
\\
\sum_{i=1}^{C_n}\delta  ( P^{n}-P_{i}^{n}  )\ast G_{\sigma^{n} } ( P^{n} )
\end{matrix}\right.
\label{eq:class}
\end{equation}
where $P_{i}^{k}$ represents the point annotation of the $i$-th instance belonging to the $k$-th class. There are $C_{k}$ instances in the $k$-th class. $\sigma^{k}$ is the kernel size for the $k$-th class in Gaussian function $G$, and the value of $\sigma^{k}$ depends on the relative sizes of the $k$-th class.

\vspace{-5pt}
\subsection{Dilated-Scale-Aware Module}\label{section3.2}

% Due to the perspective effect, the intra-scene scale variates greatly that forms challenges for feature extraction especially for multi-class counting task where the inter-class scale is not uniform.
Due to the perspective effect, the scales of objects change a lot in a scene. It forms a challenge for feature extraction. To overcome this issue, we propose the Dilated-Scale-Aware Module (DSAM) that captures multi-scale information by fusing features from different layers of the backbone. 

In the VGG16 with max-pooling layers, the receptive field of convolution layers become progressively larger as the depth increases. The intermediate layers of a CNN hierarchically learn visual patterns from edges and corners to parts and objects. As shown in Fig.~\ref{fig:pipeline}, the proposed DSAM applies different dilated convolutions on stage$\_3$, stage$\_4$ and stage$\_5$. The expansion ratios of dilated convolutional filters are set as $2$, $3$, and $5$,  respectively. After the upsample and concatenation operation, a $3\times3$ convolutional layer is added to fuse these multi-scale features. By doing so, the DSAM effectively combines hierarchical features from different layers, which indeed improves the object counting performance.

\begin{figure*}[t]
\centering
\resizebox{0.97\textwidth}{!}{
    \includegraphics{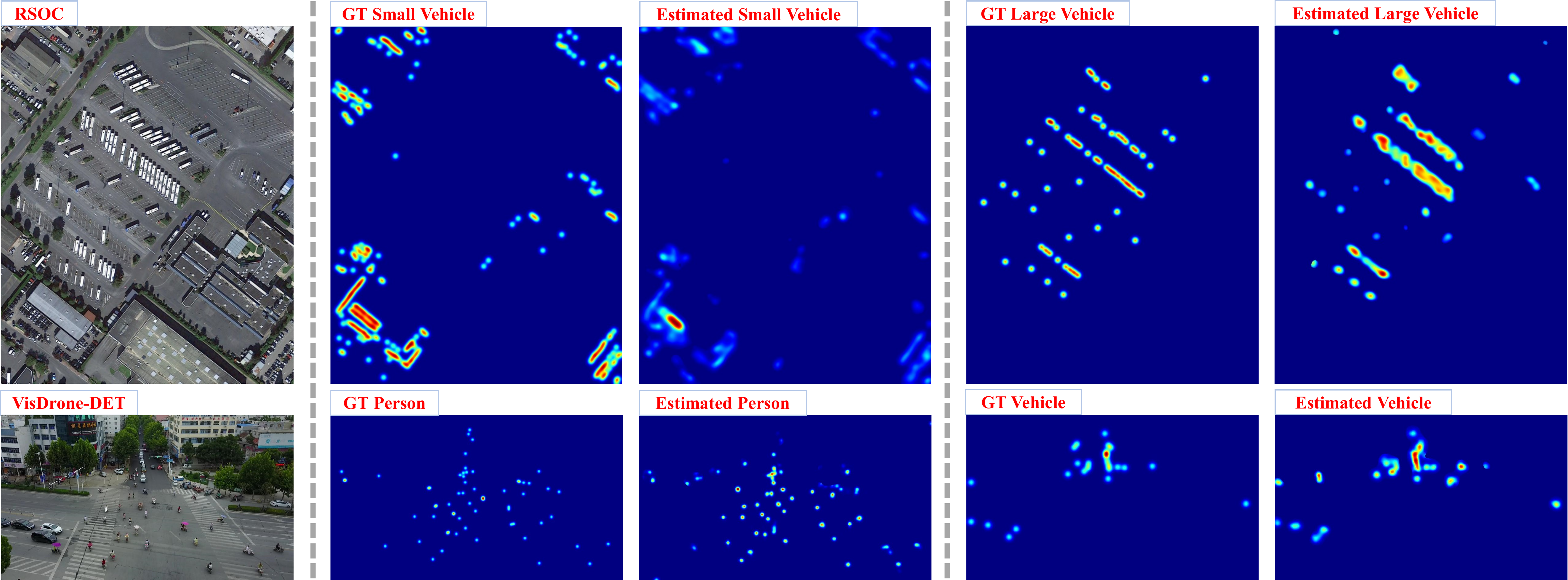}
    }
\caption{Visualization results on the RSOC dataset~\cite{gao2020counting} and the VisDrone-DET dataset~\cite{zhu2018visdrone}. The given images of the two datasets contain huge differences in viewpoint, scene, target size, and image scale, but our method shows strong robustness as well as impressive multi-class object counting performance.}
\centering
\label{fig:vision-cropped}
\vspace{-8pt}
\end{figure*}

% \begin{figure}[t]
% \centering
% \resizebox{0.47\textwidth}{!}{
%     \includegraphics{Category-Attention Module-cropped.pdf}
%     }
% \caption{The structure of Category-Attention Module.}
% \centering
% \label{fig:Category-Attention Module}
% \end{figure}

\vspace{-5pt}
\subsection{Category-Attention Module}\label{section3.3}

Although the DSAM effectively mitigates the effect of intra-class scale variation, mutual interference between classes exits on the density maps. This inspires us to design the Category-Attention Module (CAM) to generate the discriminative density maps, as shown in Fig.~\ref{fig:pipeline}.

The CAM is designed to generate spatial attentions for each class. We generate the pseudo ground-truth mask maps for each instance from point annotations to supervise the learning of spatial attentions. Specifically, we first utilize the distance transformation~\cite{rosenfeld1968distance} and transform the point annotation map to a distance map $D$, which is described as:
\begin{equation}
P_{\left ( x,y \right )} = \underset{\left ( x_{i}, y_{i} \right )\in A}{\min}\sqrt{\left ( x-x_{i} \right )^{2}+\left (y-y_{i} \right )^{2}},
\label{eq:distance_transform}
\end{equation}
where $P_{\left ( x,y \right )}$ represents the pixel value located at $\left ( x,y \right )$ in the distance map $D$. $A$ is the set of annotated coordinates. The pseudo-labels for spatial attention generation can be obtained as a $0-1$ mask by using a distance threshold $J$ to divide the distance map into the foreground and background.

As shown in Fig.~\ref{fig:pipeline}, the CAM adopts the feature maps of stage$\_5$ as input since the high-level layer learns more semantic information. Similar to the DSAM, the CAM enhances the multi-scale information via four dilated convolutional layers with $1$, $2$, $3$, $4$ dilation rates, respectively. The multi-scale feature maps are combined via channel-wise concatenation operation and several convolution layers. The number of the spatial attentions is equal to the number of categories, and the estimated spatial attentions multiplies the estimated density maps to obtain the final density maps, which effectively reduces the inter-class interference.

\vspace{-5pt}
\subsection{Loss function}
% As shown in Fig.~\ref{fig:pipeline}, there are tree outputs in our method during the training stage. Following the previous framework~\cite{zhang2016single, li2018csrnet, liu2019context}, 

We utilize the $L_2$ loss to measure the difference between the ground truth density maps and estimated density maps:
\begin{equation}
L_{2}=\sum_{n=1}^{N}\sum_{x=1}^{W}\sum_{y=1}^{H}\left | P^{'}_{(n,x,y)}-P_{(n,x,y)} \right |^{2},
\label{eq:mse}
\end{equation}
where $N$ is the total number of the categories. $W$ and $H$ represent the width and height of the training image. $ P^{'}_{(n,x,y)}$ and $ P_{(n,x,y)}$ represent the pixel values located at $(x,y)$ in the estimated density map and the ground-truth density map of the $c$-th class respectively.

We train the CAM based on the Binary Cross Entropy loss, as defined by:
\begin{equation}
\begin{aligned}
L_{BCE}=&-\frac{1}{ W \times H}\sum_{x=1}^{W}\sum_{y=1}^{H}\left ( \left ( T_{\left ( x,y \right )}\times \log{R_{\left ( x,y \right )}} \right ) \right.
\\
&\left.+\left ( 1- T_{\left ( x,y \right )}\right ) \times \log{\left ( 1-R_{\left ( x,y \right ) }\right )}  \right ),
\label{eq:ce}
\end{aligned}
\end{equation}
where $T_{\left ( x,y \right )}\in{\{0,1\}}$ denotes the pseudo-label for spatial attention located at $(x,y)$. $R_{\left ( x,y \right ) } \in{[0,1]}$ represents the predicted spatial attention. The final training objective function $L$ is defined as below:
\begin{equation}
L=\sum_{i=1}^{2}L_{2}^{i}+\sum_{n=1}^{N}L_{BCE}^{n}.
\label{eq:loss}
\end{equation}

% As described in Section~\ref{section3.3}, the Category-Attention Module achieves pixel-wise classification. And in specific, the number of output channels in the Category-Attention Module is equal to double of the number of the categories. We assign two independent channels to each category and use the Cross-Entropy (CE) loss for better supervision.
% \begin{equation}
% \mathcal{L}_{CE}(y,p)=-\frac{1}{N}\sum_{n=1}^{N}\sum_{c=1}^{2}y_{n,c}\log(p_{n,c})
% \label{eq:ce}
% \end{equation}
% Where $y\in{\{0,1\}}$ denotes ground-truth value. $p\in{[0,1]}$ represents the predicted probability. $N$ is the total number of pixels.
% The final loss function in our proposed can be written as:
% \begin{equation}
% \mathcal{L}=\sum_{i=1}^{2}\lambda_{i}\mathcal{L}_{MSE_i}+\sum_{c=1}^{C}\lambda_{c}\mathcal{L}_{CE_c}
% \label{eq:loss}
% \end{equation}
% Here $C$ is the number of the categories and $\lambda$ is set as 1 without further adjustments.  

\vspace{-5pt}
\section{Experiments}

% \vspace{-5pt}
\subsection{Implementation details}
The proposed network is trained in an end-to-end fashion. For the first 13 convolutional layers, we initialize them with a pre-trained VGG-16~\cite{simonyan2015very}. The other layers are initialized by a Gaussian initialization with 0.01 standard deviation. The distance threshold $J$ is set as $20$. We flip each image and crop the given images randomly as the data augmentation processes. The parameters are updated by an adaptive moment estimation optimizer.

\begin{figure}[t]
\centering
\resizebox{0.47\textwidth}{!}{
    \includegraphics{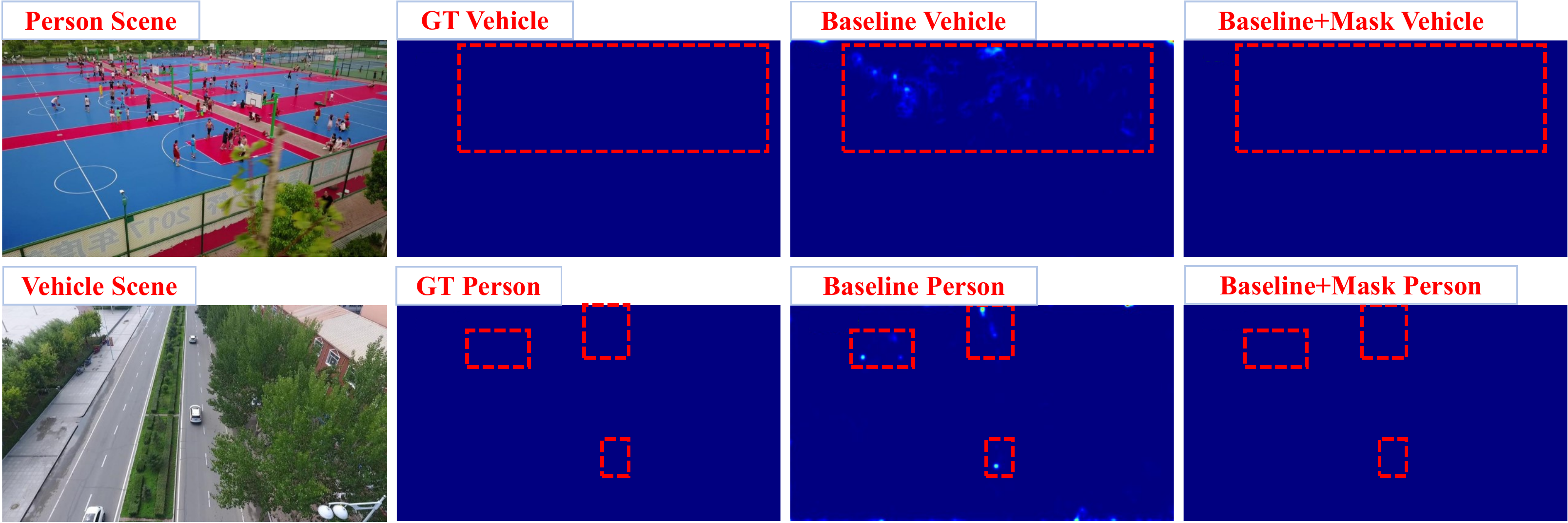}
    }
\caption{In the person-vehicle counting task, both estimated density maps are interfered by each other, as shown in the red dashed boxes. The spatial attention masks suppressed these harmful responses effectively.}
\centering
\label{fig:mask-cropped}
\vspace{-10pt}
\end{figure}

\vspace{-5pt}
\subsection{Datasets}

Since there are no publicly available multi-class object counting datasets, we utilize the single-class object counting dataset RSOC~\cite{gao2020counting} including multiple subsets and the object detection dataset VisDrone-DET~\cite{zhu2018visdrone} for evaluation. Note that object detection datasets provide box-level annotations instead of point-level annotations. We obtain the point-level annotations by computing the center of bounding boxes. All the experiments are conducted on the point-level annotations.

\textbf{VisDrone-DET Dataset~\cite{zhu2018visdrone}.} This dataset contains $10,209$ images captured by cameras equipped on drones. There are ten categories annotated, including pedestrian, person, car, van, bus, truck, motor, bicycle, awning-tricycle, and tricycle. To increase the density of the counting instances and ensure the challenge for counting task, we merge the pedestrian category into the person category and consider car, van, bus, truck as the same category named vehicle. The extremely small size of objects is the major challenge of this dataset. 

\textbf{RSOC Dataset~\cite{gao2020counting}.} This dataset contains $3,057$ images collected from Google Earth and DOTA dataset~\cite{Xia_2018_CVPR}.  It is the largest object counting dataset collected in a remote sense. And the instances included are in high density. There are four categories of images, including buildings, small vehicles, large vehicles, and ships. Specifically, the small vehicles and the large vehicles always appear in the same image, which meets the labeling requirements for multi-class object counting tasks. In our experiments, we choose the small vehicle and large vehicle subsets into a multi-class object counting dataset. The most challenging issue of the RSOC dataset is the severe background noise.

% %%%%%%%%%%%%%%%%%%%%%%%%%%%%%%%%%%%%%%%%%%%%%%%%%%%%%%%%%%%%%%%%%%%%%%%%%%%%%%%%%
% \begin{table}[t]
% \small
% \footnotesize
% \centering
% \caption{The parameter number of the proposed framework compared to other methods in a $k$-category counting task.}
% \vspace{-5pt}
% \label{tab:parameters}
% \setlength{\tabcolsep}{8mm}
% % \renewcommand\arraystretch{0.8}
% \resizebox{0.47\textwidth}{!}{
% \begin{tabular}{ l|c|c }
% \hline
%  Method &Backbone &PARAMS\\
% \hline
% MCNN~\cite{zhang2016single}&-&0.134M$ \times k$\\
% SANet~\cite{cao2018scale}&-&1.389M$ \times k$\\
% CSRNet~\cite{li2018csrnet}&VGG16&16.263M$ \times k$\\
% BL~\cite{ma2019bayesian}&VGG19&21.500M$ \times k$\\
% CAN~\cite{liu2019context}&VGG16&18.104M$ \times k$\\
% \hline
% \textbf{Ours}&VGG16&26.740M\\
%  \hline
% \end{tabular}
% }
% \vspace{-5pt}
% \end{table}
% %%%%%%%%%%%%%%%%%%%%%%%%%%%%%%%%%%%%%%%%%%%%%%%%%%%%%%%%%%%%%%%%%%%%%%%%%%%%%%%%%

%%%%%%%%%%%%%%%%%%%%%%%%%%%%%%%%%%%%%%%%%%%%%%%%%%%%%%%%%%%%%%%%%%%%%%%%%%%%%%%%%
\begin{table}[t]
\small
\footnotesize
\centering
\caption{Performance comparisons on the VisDrone-DET dataset~\cite{zhu2018visdrone}. The first and second places are marked in red and blue respectively.}
\vspace{-5pt}
\setlength{\tabcolsep}{1mm}
\resizebox{0.47\textwidth}{!}{
\begin{tabular}{ l|c|c|c|c|c|c|c }
\hline
 {\multirow{2}{*}{Method}} &{\multirow{2}{*}{Backbone}}  &\multicolumn{2}{c|}{Mean}&\multicolumn{2}{c|}{Vehicle}&\multicolumn{2}{c}{Person}\\
 \cline{3-8}
&&MAE&MSE&MAE&MSE&MAE&MSE\\
\hline
% Hydra-3s~\cite{onoro2016towards}&10.99&13.75&16.69&19.32\\
% FCN-HA~\cite{zhang2017fcn}&4.21&-&-&-\\
% CRNet~\cite{li2018csrnet}&3.56&5.49&8.75&15.04\\
% ADCrowdNet~\cite{liu2019adcrowdnet}&2.44&4.41&6.78&13.58\\
% \hline
% SCAR~\cite{gao2019scar}&16.831&24.197&18.047&23.207&15.616&25.188\\
% SFCN\dag~\cite{wang2019learning}&16.813&24.065&17.992&23.234&15.635&24.896\\
MCNN~\cite{zhang2016single}&-&9.664&14.123&8.808&11.885&10.520&16.361\\
SANet~\cite{cao2018scale}&-&16.807&24.091&17.992&23.234&15.623&24.947\\
CSRNet~\cite{li2018csrnet}&VGG16&6.778&9.429&6.022&8.091&7.535&10.767\\
BL~\cite{ma2019bayesian}&VGG19&7.503&10.504&6.754&9.021&8.252&11.988\\
CAN~\cite{liu2019context}&VGG16&\color{blue}{6.224}&\color{blue}{8.831}&\color{blue}{5.662}&\color{blue}{7.667}&\color{blue}{6.786}&\color{blue}{9.994}\\
\hline
\textbf{Ours}&VGG16&\color{red}{4.030}&\color{red}{6.285}&\color{red}{3.437}&\color{red}{5.468}&\color{red}{4.624}&\color{red}{7.102}\\
 \hline
\end{tabular}
}
\label{tab:visdrone_performance}
% \vspace{-5pt}
\end{table}
%%%%%%%%%%%%%%%%%%%%%%%%%%%%%%%%%%%%%%%%%%%%%%%%%%%%%%%%%%%%%%%%%%%%%%%%%%%%%%%%%

%%%%%%%%%%%%%%%%%%%%%%%%%%%%%%%%%%%%%%%%%%%%%%%%%%%%%%%%%%%%%%%%%%%%%%%%%%%%%%%%%
\begin{table}[t]
\small
\footnotesize
\centering
\caption{Performance comparisons on the RSOC dataset~\cite{gao2020counting}. The first and second places are marked in red and blue respectively.}
\vspace{-5pt}
\setlength{\tabcolsep}{1mm}
\resizebox{0.47\textwidth}{!}{
\begin{tabular}{ l|c|c|c|c|c|c|c }
\hline
 {\multirow{2}{*}{Method}}&{\multirow{2}{*}{Backbone}}  &\multicolumn{2}{c|}{Mean}&\multicolumn{2}{c|}{Small-Vehicle}&\multicolumn{2}{c}{Large-Vehicle}\\
 \cline{3-8}
&&MAE&MSE&MAE&MSE&MAE&MSE\\
\hline
% Hydra-3s~\cite{onoro2016towards}&10.99&13.75&16.69&19.32\\
% FCN-HA~\cite{zhang2017fcn}&4.21&-&-&-\\
% CRNet~\cite{li2018csrnet}&3.56&5.49&8.75&15.04\\
% ADCrowdNet~\cite{liu2019adcrowdnet}&2.44&4.41&6.78&13.58\\
% \hline

% SFCN\dag~\cite{wang2019learning}&&&&&&\\
% SCAR~\cite{gao2019scar}&&&&&&\\
MCNN~\cite{zhang2016single}&-&105.600&425.029&186.913&814.102&24.288&35.957\\
SANet~\cite{cao2018scale}&-&143.670&500.228&246.830&939.159&40.509&61.298\\
CSRNet~\cite{li2018csrnet}&VGG16&87.537&361.727&154.898&697.101&\color{blue}{20.177}&\color{red}{26.353}\\
BL~\cite{ma2019bayesian}&VGG19&107.212&297.607&172.883&534.460&41.541&60.754\\
CAN~\cite{liu2019context}&VGG16&\color{blue}{58.987}&\color{blue}{137.238}&\color{blue}{94.772}&\color{blue}{235.974}&23.201&38.501\\
\hline
\textbf{Ours}&VGG16&\color{red}{42.432}&\color{red}{127.441}&\color{red}{65.398}&\color{red}{223.578}&\color{red}{19.467}&\color{blue}{31.305}\\
 \hline
\end{tabular}
}
\label{tab:rsoc_performance}
\vspace{-9pt}
\end{table}
%%%%%%%%%%%%%%%%%%%%%%%%%%%%%%%%%%%%%%%%%%%%%%%%%%%%%%%%%%%%%%%%%%%%%%%%%%%%%%%%%

%%%%%%%%%%%%%%%%%%%%%%%%%%%%%%%%%%%%%%%%%%%%%%%%%%%%%%%%%%%%%%%%%%%%%%%%%%%%%%%%%
\begin{table}[t]
\small
\footnotesize
\centering
\caption{Ablation studies on the VisDrone-DET dataset~\cite{zhu2018visdrone}. The first and second places are marked in red and blue respectively.}
\vspace{-5pt}
\setlength{\tabcolsep}{1mm}
\resizebox{0.47\textwidth}{!}{
\begin{tabular}{ l|c|c|c|c|c|c }
\hline
 {\multirow{2}{*}{Method}}  &\multicolumn{2}{c|}{Mean}&\multicolumn{2}{c|}{Vehicle}&\multicolumn{2}{c}{Person}\\
 \cline{2-7}
&MAE&MSE&MAE&MSE&MAE&MSE\\
\hline
% Hydra-3s~\cite{onoro2016towards}&10.99&13.75&16.69&19.32\\
% FCN-HA~\cite{zhang2017fcn}&4.21&-&-&-\\
% CRNet~\cite{li2018csrnet}&3.56&5.49&8.75&15.04\\
% ADCrowdNet~\cite{liu2019adcrowdnet}&2.44&4.41&6.78&13.58\\
% \hline
baseline&6.823&9.549&6.242&8.363&7.405&10.736\\
baseline+DSAM&5.066&7.439&4.233&6.049&5.900&8.828\\
baseline+CAM&\color{blue}{4.202}&\color{blue}{6.649}&\color{blue}{3.556}&\color{red}{5.455}&\color{blue}{4.849}&\color{blue}{7.844}\\
baseline+DSAM+CAM&\color{red}{4.030}&\color{red}{6.285}&\color{red}{3.437}&\color{blue}{5.468}&\color{red}{4.624}&\color{red}{7.102}\\
 \hline
\end{tabular}
}
\label{tab:ablation_study}
\vspace{-11pt}
\end{table}
%%%%%%%%%%%%%%%%%%%%%%%%%%%%%%%%%%%%%%%%%%%%%%%%%%%%%%%%%%%%%%%%%%%%%%%%%%%%%%%%%

\vspace{-5pt}
\subsection{Evaluation metrics}
The Mean Absolute Error (MAE) and the Mean Squared Error (MSE) are the primary metrics in the object counting task. In the multi-class counting task, we adopt MAE and MSE to evaluate counting performance for each class, then average the results of all the categories to get $MAE_{Mean}$ and $MSE_{Mean}$, which denote the multi-class counting capability:
\begin{equation}
MAE=\frac{1}{M}\sum_{m=1}^{M}\left | Q_m^{'}-Q_m \right |, MSE=\sqrt{\frac{1}{M}\sum_{m=1}^{M}\left ( Q^{'}_m-Q_m \right )^2},
\label{eq:metric}
\end{equation}
\begin{equation}
MAE_{Mean}=\frac{1}{N}\sum_{n=1}^{N}MAE_n, MSE_{Mean}=\frac{1}{N}\sum_{n=1}^{N}MSE_n
\label{eq:Overall-metric},
\end{equation}
where $M$ denotes the number of images in the testing set, $N$ is the number of the categories. $Q_m^{'}$ and $Q_m$ are the estimated and labeling count number of the $m$-th image.

% For VisDrone-DET dataset~\cite{zhu2018visdrone}, there were 100 epochs in total. The learning rate was set as $10^{-5}$ and then divided by ten at the 70th and 90th epoch. The given images were cropped to 384$\times$384 pixels at random locations. 

% For RSOC dataset~\cite{gao2020counting}, there were 300 epochs in total. The learning rate was initialized to be $10^{-5}$ and decayed tenfold at the 250th and 280th epoch. The given images were cropped to 512$\times$512 pixels at random locations.

% \section{Results and Analysis}

\vspace{-5pt}
\subsection{Comparisons with the State-of-the-art}
The traditional object counting methods are designed for single-class object counting task. While for multi-class object counting task, the previous point-level annotations based object counting methods should train separate models for each category. 
% We list the number of parameters for each method in Tab.~\ref{tab:parameters}. Our method requires less computational effort than traditional methods when deal with the multi-class object counting task $\left ( k\geq 2 \right )$.
The experiments on these two datasets are two-category object counting. For fair comparisons, we change the output channel of other methods to $2$ to implement the multi-class object counting task. Tab.~\ref{tab:visdrone_performance} and Tab.~\ref{tab:rsoc_performance} list the quantitative results of VisDrone-DET~\cite{zhu2018visdrone} and RSOC~\cite{gao2020counting} respectively, and we also tell the quality of estimated density maps in Fig.~\ref{fig:vision-cropped}.

\textbf{VisDrone-DET Dataset.} We first compare our method with various classic object counting methods on the VisDrone-DET dataset. The proposed method achieves the best performance on all metrics. Specially, our method improves $35.3\%$ in $MAE_{Mean}$, $28.8\%$ in $MSE_{Mean}$, $39.3\%$ in $MAE_{Vehicle}$, $28.7\%$ in $MSE_{Vehicle}$, $31.9\%$ in $MAE_{Person}$ and $28.9\%$ in $MSE_{Person}$ compared with CAN.

\textbf{RSOC Dataset.} We also compare our method with other related works on the RSOC dataset. As we can see, the proposed method achieves $42.432$ in $MAE_{Mean}$ and $127.441$ in $MSE_{Mean}$, which is currently the best multi-class object counting performance. Counting the Small-Vehicle, our approach improves $31.0\%$ in $MAE_{Small-Vehicle}$ and achieves $12.396$ $MSE_{Small-Vehicle}$ dropped compared with CAN, respectively.

In summary, the proposed method shows outstanding performance on these two datasets.

\vspace{-5pt}
\subsection{Ablation Studies}
In this section, we conduct the ablation studies to verify the effectiveness of DSAM and CAM on the VisDrone-DET dataset~\cite{zhu2018visdrone}.

\textbf{Effectiveness of DSAM.} We first study the impact of DSAM, which is designed to capture the intra-class multi-scale information. The results are listed in Tab.~\ref{tab:ablation_study}. We find that the addition of DSAM causes $1.757$ $MAE_{Mean}$ and $2.11$ $MSE_{Mean}$ dropped compared with the baseline.

\textbf{Effectiveness of CAM.} We further validate the effect of the proposed CAM. The results are listed in Tab.~\ref{tab:ablation_study}. CAM achieves $38.4\%$ $MAE_{Mean}$ and $30.4\%$ $MSE_{Mean}$ improvement, which indicates that CAM effectively reduces the inter-class interference on density maps. By observing Fig.~\ref{fig:mask-cropped}, we can see that the predicted results with CAM exhibit better clarity and better accuracy compared with baseline in both crowd scenes and vehicle scenes.

% %%%%%%%%%%%%%%%%%%%%%%%%%%%%%%%%%%%%%%%%%%%%%%%%%%%%%%%%%%%%%%%%%%%%%%%%%%%%%%%%%
% \begin{table}[htbp]
% \small
% \footnotesize
% \centering
% \setlength{\tabcolsep}{2.5mm}
% % \renewcommand\arraystretch{0.8}
% \resizebox{0.47\textwidth}{!}{
% \begin{tabular}{ l|c|c|c|c|c|c }
% \hline
%  {\multirow{2}{*}{Method}}  &\multicolumn{2}{c|}{Overall}&\multicolumn{2}{c|}{Car}&\multicolumn{2}{c}{Person}\\
%  \cline{2-7}
% &MAE&MSE&MAE&MSE&MAE&MSE\\
% \hline
% % Hydra-3s~\cite{onoro2016towards}&10.99&13.75&16.69&19.32\\
% % FCN-HA~\cite{zhang2017fcn}&4.21&-&-&-\\
% % CRNet~\cite{li2018csrnet}&3.56&5.49&8.75&15.04\\
% % ADCrowdNet~\cite{liu2019adcrowdnet}&2.44&4.41&6.78&13.58\\
% % \hline
% CSRNet~\cite{li2018csrnet}&&&&&&\\
% CAN~\cite{liu2019context}&&&&&&\\
%  \whline
% \textbf{Ours} &4.030&6.285&3.437&5.468&4.624&7.102\\
%  \hline
% \end{tabular}
% }
% \caption{Single-class vs. Multi-class VisDrone2019 dataset}
% \label{tab:vehicle}
% \vspace{-5pt}
% \end{table}
% %%%%%%%%%%%%%%%%%%%%%%%%%%%%%%%%%%%%%%%%%%%%%%%%%%%%%%%%%%%%%%%%%%%%%%%%%%%%%%%%%

\section{Conclusion}
In this paper, we propose a novel Dilated-Scale-Aware Category-Attention ConvNet (DSACA), which achieves multi-class object counting simultaneously only based on point-level annotations. To overcome the inter-class interference that appears among the density maps and mitigate the effects of intra-class scale variations, we designed Dilated-Scale-Aware Module (DSAM) and Category-Attention Module (CAM). Extensive experiments on two challenging benchmarks show the validity of our approach. In the future work, we will explore the correlation of multi-class object counting and single-class object counting systematically.

% List and number all bibliographical references at the end of the paper. The references can be numbered in alphabetic order or in order of appearance in the document. When referring to them in the text, type the corresponding reference number in square brackets as shown at the end of this sentence~\cite{Morgan2005}. All citations must be adhered to IEEE format and style. Examples such as~\cite{Morgan2005},~\cite{cooley65} and~\cite{haykin02} are given in Section 12.


% \clearpage

\bibliographystyle{ieeetr}
\bibliography{reference}

\end{document}